\pdfoutput=1
\documentclass[letterpaper, 10 pt, conference]{ieeeconf}
\usepackage{graphicx}
\usepackage{amsmath}
\usepackage{amssymb}
\usepackage{threeparttable}
\usepackage{booktabs}
\usepackage{color}
\usepackage{algorithm}
\usepackage{algorithmic}
\usepackage{pifont}
\usepackage{mathrsfs}
\usepackage{multirow}
\usepackage{threeparttable}
\usepackage{flushend}
\usepackage{makecell}
\usepackage{bm}
\usepackage{tabularx}
\usepackage{csquotes}
\let\labelindent\relax\usepackage{enumitem}
\usepackage[colorlinks,linkcolor=blue]{hyperref}
\setlist[itemize]{leftmargin=*}
\newcommand{\cmark}{\ding{51}}
\newcommand{\xmark}{\ding{55}}

\newcommand{\vect}[1]{\boldsymbol{\mathbf{#1}}}

\IEEEoverridecommandlockouts
\begin{document}

\title{\LARGE \bf Towards Physically Realizable Adversarial Attacks in \\ Embodied Vision Navigation}

\author{Meng Chen$^{1}$,\ Jiawei Tu$^{1}$,\ Chao Qi$^{1}$,\ Yonghao Dang$^{1}$,\ Feng Zhou$^{1}$,\ Wei Wei$^{1}$,\ Jianqin Yin$^{1,*}$\
\thanks{$^{1}$School of Intelligent Engineering and Automation, Beijing University of Posts and Telecommunications, Beijing 100876, China.}
\thanks{$^{*}$Corresponding author.}
\thanks{e-mail: chenmeng@bupt.edu.cn.}
}

\markboth{2024 IEEE International Conference on Robotics and Automation (ICRA 2024)}%
{Shell \MakeLowercase{\textit{et al.}}: Bare Demo of IEEEtran.cls for IEEE Journals}

\maketitle
\pagestyle{empty}
\thispagestyle{empty}

\begin{abstract} 
The significant advancements in embodied vision navigation have raised concerns about its susceptibility to adversarial attacks exploiting deep neural networks. Investigating the adversarial robustness of embodied vision navigation is crucial, especially given the threat of 3D physical attacks that could pose risks to human safety. However, existing attack methods for embodied vision navigation often lack physical feasibility due to challenges in transferring digital perturbations into the physical world. Moreover, current physical attacks for object detection struggle to achieve both multi-view effectiveness and visual naturalness in navigation scenarios. To address this, we propose a practical attack method for embodied navigation by attaching adversarial patches to objects, where both opacity and textures are learnable. Specifically, to ensure effectiveness across varying viewpoints, we employ a multi-view optimization strategy based on object-aware sampling, which optimizes the patch's texture based on feedback from the vision-based perception model used in navigation. To make the patch inconspicuous to human observers, we introduce a two-stage opacity optimization mechanism, in which opacity is fine-tuned after texture optimization. Experimental results demonstrate that our adversarial patches decrease the navigation success rate by an average of 22.39\%, outperforming previous methods in practicality, effectiveness, and naturalness. Code is available at: \href{https://github.com/chen37058/Physical-Attacks-in-Embodied-Nav}{github.com/chen37058/Physical-Attacks-in-Embodied-Nav}.
\end{abstract}

\IEEEpeerreviewmaketitle

\section{Introduction}\label{sec:introduction}

\begin{figure}[t!]
	\centering
	\includegraphics[width=0.99\linewidth]{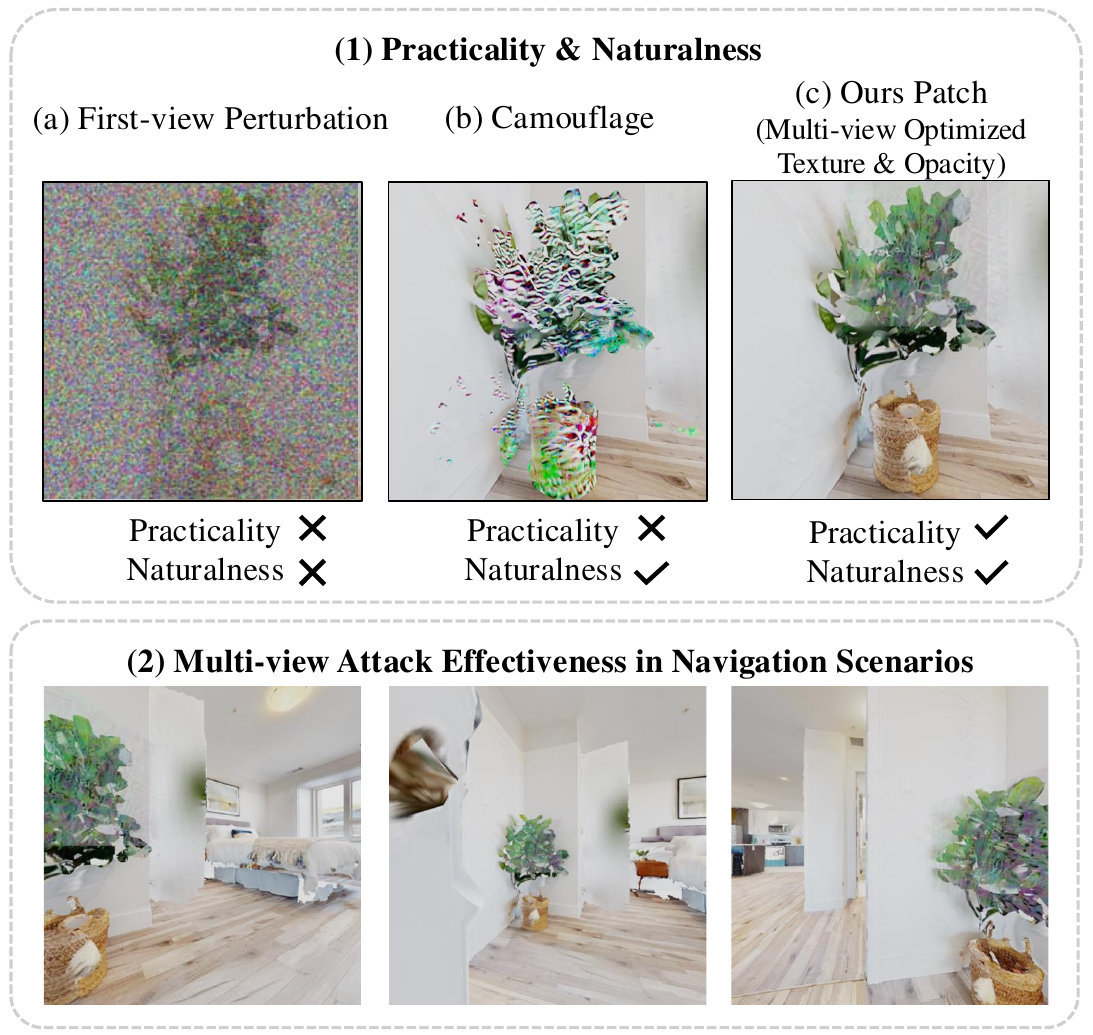}
	\caption{\textbf{Motivation.} Previous attacks were either impractical or failed to achieve both naturalness and multi-view effectiveness in navigation. For instance, prior methods such as (a) applied universal perturbations to first-person views, (b) altered 3D object textures for camouflage. In contrast, our patch (c), with learnable textures and opacity, appears more natural and, as shown in the lower panel (2), effectively prevents detection from multiple viewpoints when placed on the target object.}
	\label{fig:motivation}
	\vspace{-0.1cm}
\end{figure}

Embodied vision navigation~\cite{chaplot2020object,wang2022towards} refers to an agent moving towards a target object or a designated location in an unseen environment. It is widely applied in safety-critical scenarios, such as aiding individuals with disabilities in finding objects within their homes. The integration of Deep Neural Networks (DNNs) enables embodied navigation agents to leverage vision-based signal processing and sequential decision-making techniques~\cite{zhu2021deep}, driving significant advancements. However, DNNs are known to be vulnerable to adversarial examples~\cite{goodfellow2014explaining, carlini2017towards, xiao2018generating}, posing security risks for DNN-based embodied navigation systems in real-world deployments. In high-risk scenarios such as assistive robotics, security patrolling, and industrial automation, adversarial attackers could exploit these vulnerabilities to manipulate the agent’s decisions, leading to incorrect path planning, target deviations, or even direct threats to human safety. Therefore, studying adversarial attacks on embodied navigation—particularly 3D physical adversarial attacks—is crucial for identifying adversarial robustness issues and advancing the theoretical and technological foundations of secure navigation systems.

Recently, research on adversarial attacks against embodied vision navigation has been scarce. For instance, as shown in Figure~\ref{fig:motivation} (a), Ying et al.~\cite{ying2023consistent} applied universal perturbations to the agent's first-person observations. Another approach, as shown in Figure~\ref{fig:motivation} (b), Liu et al.~\cite{liu2020spatiotemporal} explored 3D adversarial camouflage by perturbing object textures in navigation scenes. Similar object-level texture perturbation attacks targeting object detectors have also been studied in~\cite{huang2024towards, suryanto2023active, suryanto2022dta, wang2022fca, wang2021dual}. However, these methods face practical challenges. Applying adversarial stickers directly to the agent's camera view is impractical, as attackers typically lack control over the agent's camera. Similarly, modifying object textures and shapes in the scene is costly, often requiring 3D printing, object replacement, or texture projection using projectors.

Other studies on physical adversarial attacks in robotics~\cite{zhu2021adversarial, alharthi2024physical} investigate the placement of small objects in the physical environment to deceive point cloud semantic segmentation in autonomous driving or grasping networks. While these attacks are effective against robots that rely solely on depth sensor information (e.g., point clouds or depth maps), they remain ineffective against navigation robots that are equipped with only RGB sensors. Several adversarial patch attacks that are physically realizable~\cite{brown2017adversarial, eykholt2018robust, yang2020patchattack, liu2018dpatch, yang2022controllable, kang2024diffender} have been widely applied in areas such as traffic sign detection~\cite{eykholt2018robust, wei2022simultaneously} and facial recognition~\cite{sharif2016accessorize}. These methods typically rely on techniques such as Expectation over Transformation (EoT)~\cite{athalye2018synthesizing, wiyatno2019physical} to model real-world transformations, including scaling and rotation. However, these methods, including the single-pixel attack proposed in the physical adversarial attack work on robotics~\cite{alharthi2024physical}, are insufficient to effectively address complex viewpoint variations in navigation scenarios. While multi-view adversarial patches~\cite{oslund2022multiview, hu2023physically} are specifically designed to address viewpoint variations, they are primarily developed for adversarial clothing. Their unnatural appearance makes them unsuitable for objects in navigation scenes, as they can be easily noticed by humans. \textbf{In summary, existing physical attack methods fail to simultaneously satisfy two key criteria in navigation tasks: (1) Insufficient attack effectiveness under complex viewpoint variations; (2) Lack of naturalness when applied to objects, making them easily detectable and easy to defend against.}

To develop a physically feasible attack method that satisfies both multi-view effectiveness and naturalness in embodied navigation scenarios, we propose attaching an adversarial patch with learnable textures and opacity to navigation target objects. \textbf{First, to enhance multi-view attack effectiveness, we employ a multi-view optimization strategy based on object-aware sampling to refine the adversarial patch.} This process selects the most informative viewpoints based on feedback from the navigation vision model. A physics-based differentiable renderer generates first-person images, which are fed into the model to compute detection loss. Gradients then optimize the patch’s texture and opacity. Figure~\ref{fig:motivation} (lower panel) illustrates its effectiveness. \textbf{Second, to improve naturalness, we incorporate an opacity optimization mechanism.} The patch comprises an RGB texture and a single-channel opacity mask, both refined via multi-view optimization, ensuring adjustable transparency and minimal perceptibility to human observers.

We conducted experiments on a fundamental embodied navigation task, ObjectNav, using the HM3D scene datasets. We compared our method with prior attack approaches, including full-coverage texture attacks on 3D objects and existing 2D adversarial pixel patches. The results show that our adversarial patches lead to an average reduction of 22.39\% in navigation success rates. 

The contributions of this research are as follows: 
\begin{itemize} 
    \setlength{\itemsep}{0pt}
    \item We propose a physically realizable attack method for embodied vision navigation by attaching adversarial patches with learnable opacity and textures to objects.
    \item We introduce a multi-view optimization using object-aware sampling to enhance patch effectiveness and opacity optimization for visual naturalness.
    \item Experimental results show that our adversarial patches outperform previous methods in practicality, effectiveness, and naturalness.
\end{itemize}

\section{Related Work}\label{sec:relatedwork}
\subsection{Embodied Vision Navigation.}
Embodied navigation~\cite{wang2022towards,gervet2023navigating} involves agents navigating unseen environments using visual inputs. Key tasks include object goal navigation~\cite{du2021vtnet,chaplot2020object,majumdar2022zson}, image goal navigation~\cite{kim2023topological}, visual language navigation~\cite{park2023visual,krantz2020beyond}, and embodied question answering~\cite{das2018embodied}. Object goal navigation, which requires locating specific object categories in unknown environments, tests scene understanding and memory, with applications like assisting individuals with disabilities. We focus on adversarial attacks in this task, which are fundamental and critical to other embodied tasks.

\subsection{Physical Adversarial Attacks.}

Early adversarial attack research on embodied navigation~\cite{liu2020spatiotemporal} focused on modifying object properties (e.g., 3D shape, texture) in key scene views. Universal perturbations~\cite{ying2023consistent} explored patch-based sensor attacks but proved impractical for real-world deployment. Recent work~\cite{he2024everyday} investigates backdoor attacks, where triggers implanted in the training data cause predefined abnormal behaviors in navigation models. This is effective when attackers can distribute compromised models to victims. In contrast, we consider a different threat model: attackers cannot alter the victim's model but can modify the physical environment, a scenario particularly relevant for protecting key assets in military applications.

Existing studies on physical adversarial attacks on robots~\cite{zhu2021adversarial} use small objects to deceive point cloud segmentation in autonomous driving, which is effective only for depth-based sensors. Similarly, ~\cite{alharthi2024physical} disrupts grasping success with a small sphere but is also limited to depth sensors and fails against navigation robots equipped only with RGB sensors. Moreover, its single-pixel attack struggles with viewpoint variations in navigation.

Several physically realizable adversarial patch attacks~\cite{brown2017adversarial, eykholt2018robust, yang2020patchattack, liu2018dpatch, yang2022controllable,wei2023distributional, kang2024diffender,xiao2021improving} have been widely applied in traffic sign detection~\cite{eykholt2018robust, wei2022simultaneously} and facial recognition~\cite{sharif2016accessorize}, modify 2D pixel space without altering the object itself. While techniques like Expectation over Transformation (EoT)~\cite{athalye2018synthesizing, wiyatno2019physical} can simulate real-world transformations (e.g., scaling, rotation), navigation scenarios involve far more diverse and unpredictable viewpoint changes, limiting attack efficacy. Multi-view adversarial patches~\cite{oslund2022multiview, hu2023physically} are mainly designed for clothing, but their unnatural appearance makes them unsuitable for objects in navigation tasks.

To address these limitations, we generate physically realizable adversarial examples for embodied navigation that ensure both multi-view effectiveness and natural appearance in the 3D physical world.

\section{Methodology}\label{sec:methodology}

\begin{figure*}[t!]
	\centering
	\includegraphics[width=0.95\linewidth]{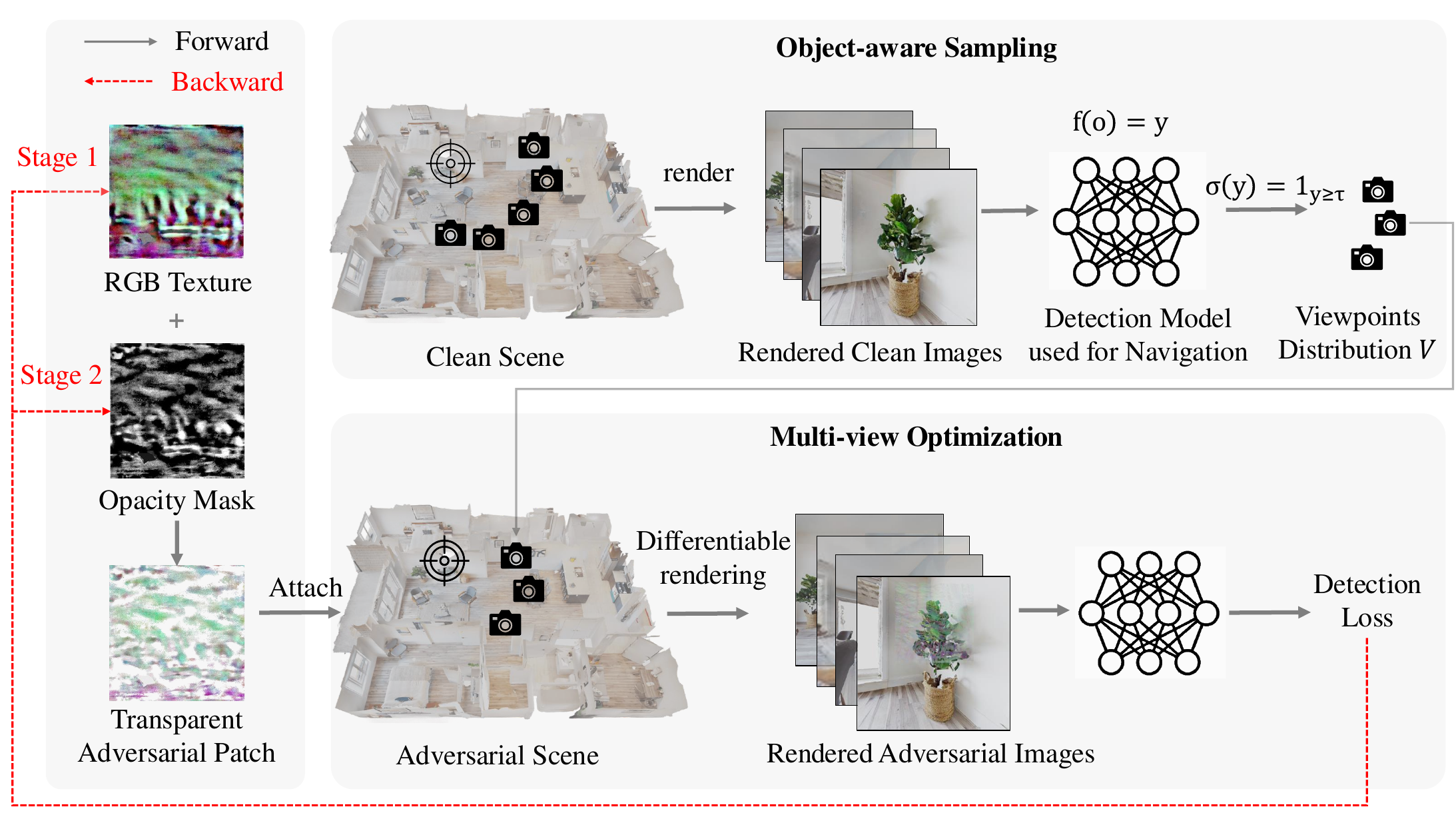}
	\caption{Method Overview. The initialized patch $\mathcal{P}^{adv}$ is attached to the target object in scene $\mathcal{S}$. (1) To enhance naturalness, the patch comprises a base texture (three RGB channels) and a single-channel opacity mask, both refined via multi-view optimization (Sec.~\ref{sec:opacity}), enabling adjustable transparency and reducing perceptibility. (2) To ensure multi-view effectiveness, we employ a multi-view optimization strategy based on object-aware sampling. Camera viewpoints are initialized in batches and filtered using the detection/segmentation model $f$, yielding an optimized viewpoint distribution $V$ (Sec.~\ref{sec:sample}). (3) We refine the patch using $V$ in multi-view optimization, leveraging a differentiable renderer $\mathcal{R}$ to generate first-person images, which are fed into $f$ to compute detection loss. Gradients are then used to optimize the patch's texture and opacity via the PGD method (Sec.~\ref{sec:optimization}).}
	\label{fig:method}
\end{figure*}

In this section, we present our method for generating adversarial patches with learnable textures and opacity, as illustrated in Figure~\ref{fig:method}.

\subsection{Problem Formulation}\label{sec:problem}
Given a scenario \(\mathcal{S}\) and a detection/segmentation model used for navigation, represented as \(f: o \rightarrow y\), our objective is to generate adversarial patches \(\mathcal{P}^{\text{adv}}\) for target objects in the scene to make these objects undetectable and cause navigation failure. Here, \(f\) denotes the detection model, \(o\) the input image, and \(y\) the model’s predicted output (e.g., bounding boxes or segmentation masks). We denote by \(\mathcal{R}(\cdot)\) a differentiable renderer that simulates the rendered image under specified lighting \(L\), patch \(\mathcal{P}^{\text{adv}}\), and viewpoint \(\vect{v}\).

To attack the detector \(f\), we define the adversarial objective as follows:
\begin{equation}\label{eq:problem}
\max_{\mathcal{P}^{\text{adv}}}\Big\{
\mathbb{E}_{\vect{v} \in V}
\mathcal{L}_{\text{attack}}\big(
f\big(\mathcal{R}(\mathcal{S}, \mathcal{P}^{\text{adv}}; L, \vect{v})\big),\, y
\big)\Big\},
\end{equation}

where \(\mathcal{L}_{\text{attack}}(\cdot)\) denotes the detection loss computed from the Mask R-CNN model. We define:
\begin{equation}\label{eq:l_attack}
\mathcal{L}_{\text{attack}} = \sum_{i=1}^{4} \lambda_i \cdot \mathcal{L}_i,
\end{equation}

where \(\mathcal{L}_i\) corresponds to the classification loss, bounding box regression loss, RPN classification loss, and RPN localization loss, respectively. The weights \(\lambda = [0.7, 0.1, 0.1, 0.1]\) reflect their relative importance during optimization.

The challenge lies in optimizing the patch to generalize across diverse viewpoints \(\vect{v}\in V\), which we address via object-aware sampling (Sec.~\ref{sec:sample}) and multi-view optimization (Sec.~\ref{sec:optimization}).

\subsection{Object-aware Sampling}\label{sec:sample}

Since target object visibility varies with angles, distances, and scene contexts, we employ a specialized sampling strategy to obtain the most valuable viewpoints for optimizing the patch. We randomly initialize camera viewpoints and filter configurations based on the detection model’s detection confidence, yielding a distribution \( V \) that is optimal for optimization. Cameras are positioned around the target object, and the viewpoint distribution is defined as:

\begin{equation}\label{eq:sample}
\begin{aligned}
V &= \left\{ (\mathbf{p}_i(r), \mathbf{\phi}_i) \mid r \in R,\, \sigma(f(o) = 1) \right\}, \\
i &\in \{0, 1, \dots, N-1\}, \\
\mathbf{p}_i(r) &= \left( c_x + r \cos \left(\frac{2\pi i}{N}\right),\, 
c_y,\, 
c_z + r \sin \left(\frac{2\pi i}{N}\right) \right), \\
\mathbf{\phi}_i &= \left( 0,\, \frac{2\pi i}{N} - \frac{\pi}{2},\, \pi \right).
\end{aligned}
\end{equation}

Here, \( \mathbf{p}_i(r) \) represents the camera position around the object center \((c_x, c_y, c_z)\), with radii \(r \in R\) ensuring layered coverage. \( \mathbf{\phi}_i \) denotes the camera orientation, and \( N \) is the number of cameras per circle. The indicator function \(\sigma(x) = 1_{x \geq \tau}\) filters configurations based on detection confidence.

\subsection{Multi-view Optimization}\label{sec:optimization}

To address the challenge of the patch’s multi-view effectiveness, we optimize the patch using a physics-based differentiable renderer and the sampled viewpoint distribution. First-person images from these viewpoints are rendered and fed into the detection model to compute the total detection loss. Gradients from multiple views are used to optimize the patch texture with a gradient-based method. The rendering process is defined as:

\begin{align}\label{eq:render}
    o &= \mathcal{R}(\mathcal{S}; L, \mathbf{p}_i(r), \mathbf{\phi}_i), \\
    \hat{o} &= \mathcal{R}(\mathcal{S}, \mathcal{P}^{adv}; L, \mathbf{p}_i(r), \mathbf{\phi}_i),
\end{align}

where \( o \) and \( \hat{o} \) are images without and with the adversarial patch, respectively. The final gradient-based update for the patch texture is:

\begin{gather}\label{eq:gradient}
\mathcal{P}_{\text{adv}} = \mathcal{P}_{\text{adv}} + \alpha \cdot \text{sign}(\nabla_{\mathcal{P}_{\text{adv}}} L_{\text{total}}(f(\mathcal{R}(\mathcal{S}_{\text{adv}}, \bm{c})), y)),
\end{gather}

where \( \alpha \) is the learning rate, and \(\nabla_{\mathcal{P}_{\text{adv}}} L_{\text{total}}\) is the gradient of the total loss with respect to the patch texture. This iterative process ensures robust adversarial patches across multiple viewpoints and lighting conditions.

\subsection{Opacity Optimization}\label{sec:opacity}
To improve the naturalness of the patch, we incorporate an opacity optimization mechanism. Digitally, our adversarial patch consists of a base texture image with three RGB color channels and a single-channel opacity mask, where transparency values range from 0 to 255, with 255 representing full opacity and 0 representing full transparency. Both the base texture and the opacity mask are optimized for effectiveness.

Intuitively, we consider the primary role of the texture to be enhancing the attack’s efficacy, while the opacity mainly serves to reduce the patch’s visibility to the human eye. Jointly optimizing both components may lead to conflicting gradient directions, especially in the early training stage, where strong textures can be paired with suboptimal opacity, resulting in visually conspicuous patches that undermine stealth.

\begin{figure}[h!]
	\centering
	\includegraphics[width=0.99\linewidth]{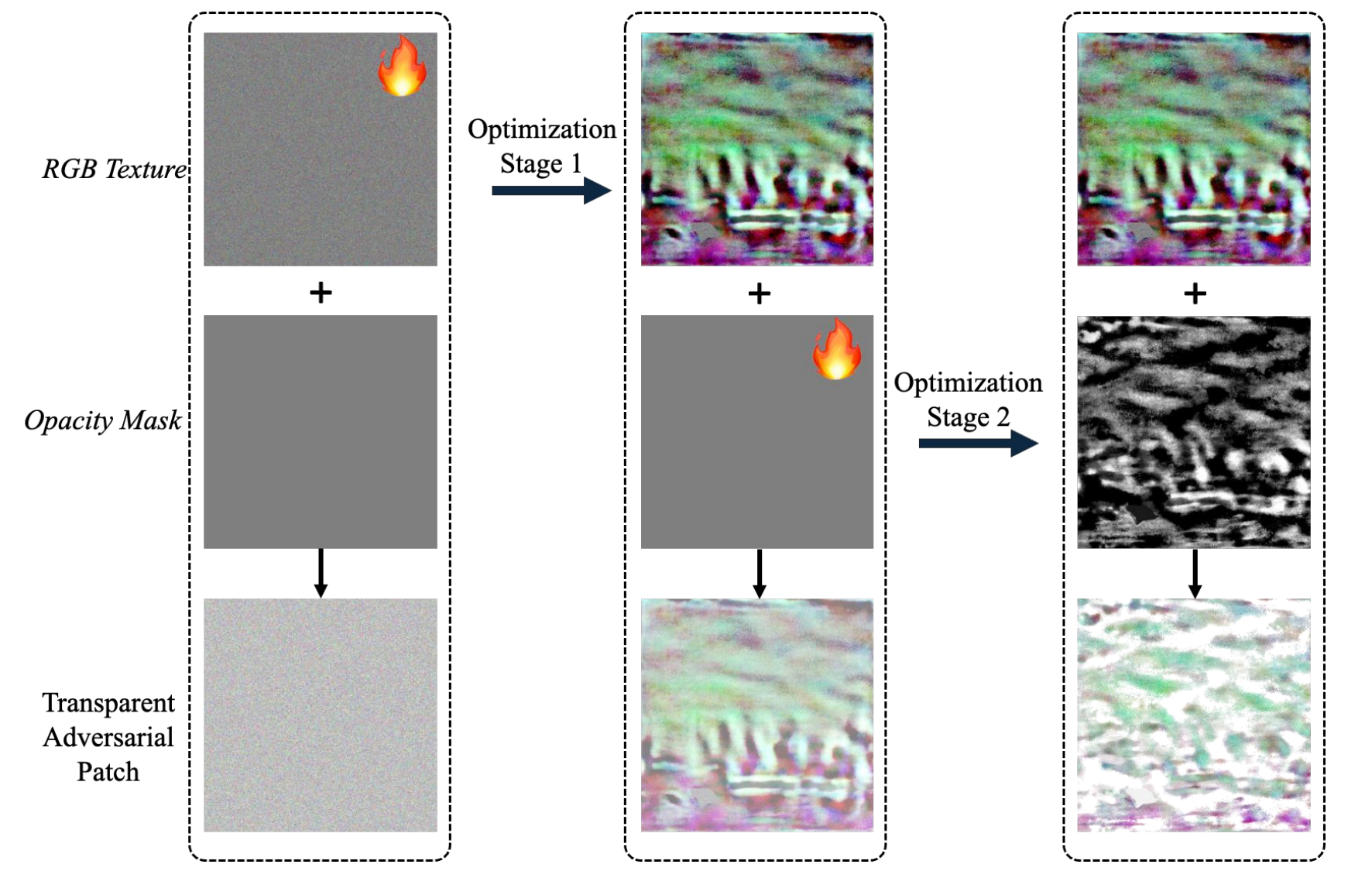}
	\caption{\textbf{Opacity Optimization Strategy.} The first stage optimizes the RGB texture to enhance the attack’s efficacy, while the second stage refines the opacity mask to reduce the patch’s detectability to the human eye.}
	\label{fig:opacity}
\end{figure}

Thus, as shown in Figure~\ref{fig:opacity}, we employ a two-stage optimization strategy. In the first stage, we randomly initialize a three-channel RGB texture and a constant single-channel opacity mask, combining them to form the initial adversarial patch. This patch is iteratively optimized using the strategies described in Section~\ref{sec:sample} and Section~\ref{sec:optimization}, resulting in an effective adversarial texture. In the second stage, we optimize the opacity mask based on the optimized texture, using the same optimization approach. All optimizations are conducted using a gradient-based PGD (Projected Gradient Descent) method.

\section{Experiments}\label{sec:exp}

\begin{table}[t]
	\centering
	\scriptsize
	\caption{Attack performance against multiple targets in HM3D for navigation}
	\label{tab:main1}
	\begin{tabular}{c|c|lll}
		\toprule
		Target & Attack Method & SR$\downarrow$(\%) & SPL$\downarrow$(\%) & DTS$\uparrow$(m)  \\
		\midrule
		\multirow{3}{*}{\shortstack{tv \\ monitor}} 
        & No Attack &  100.00 & 63.03 & 0.04   \\
		& Camouflage &  100.00\textcolor{red}{\scriptsize{(0.00$\downarrow$)}} & 56.37\textcolor{red}{\scriptsize{(6.66$\downarrow$)}}  & 0.04\textcolor{red}{\scriptsize{(0.00$\uparrow$)}} \\
		& \textit{Ours} Patch &  \textbf{60.00} \textcolor{red}{\scriptsize{(40.00$\downarrow$)}} & \textbf{14.81} \textcolor{red}{\scriptsize{(48.22$\downarrow$)}}  & \textbf{1.68}\textcolor{red}{\scriptsize{(1.64$\uparrow$)}} \\
		\midrule
        \multirow{3}{*}{plant} 
        & No Attack &  93.75 & 26.60 & 0.24   \\
		& Camouflage &  \textbf{37.50}\textcolor{red}{\scriptsize{(56.25$\downarrow$)}} & 11.87\textcolor{red}{\scriptsize{(14.73$\downarrow$)}}  & \textbf{1.06}\textcolor{red}{\scriptsize{(0.82$\uparrow$)}} \\
		& \textit{Ours} Patch &  62.50 \textcolor{red}{\scriptsize{(31.25$\downarrow$)}} & \textbf{7.98} \textcolor{red}{\scriptsize{(18.62$\downarrow$)}}  & 0.73\textcolor{red}{\scriptsize{(0.49$\uparrow$)}} \\
        \midrule
        \multirow{3}{*}{bed} 
        & No Attack &  94.11 & 55.83 & 0.16   \\
		& Camouflage &  94.11\textcolor{red}{\scriptsize{(0.00$\downarrow$)}} & 55.95\textcolor{red}{\scriptsize{(0.12$\downarrow$)}}  & 0.16\textcolor{red}{\scriptsize{(0.00$\uparrow$)}} \\
		& \textit{Ours} Patch &  \textbf{94.11} \textcolor{red}{\scriptsize{(0.00$\downarrow$)}} & \textbf{46.44} \textcolor{red}{\scriptsize{(9.39$\downarrow$)}}  & \textbf{0.16}\textcolor{red}{\scriptsize{(0.00$\uparrow$)}} \\
		\midrule
        \multirow{3}{*}{chair} 
        & No Attack &  100.00 & 65.18 & 0.22   \\
		& Camouflage &  100.00\textcolor{red}{\scriptsize{(0.00$\downarrow$)}} & 58.97\textcolor{red}{\scriptsize{(6.21$\downarrow$)}}  & 0.22\textcolor{red}{\scriptsize{(0.00$\uparrow$)}} \\
		& \textit{Ours} Patch &  \textbf{83.33} \textcolor{red}{\scriptsize{(16.67$\downarrow$)}} & \textbf{47.12} \textcolor{red}{\scriptsize{(18.06$\downarrow$)}}  & \textbf{0.34}\textcolor{red}{\scriptsize{(0.12$\uparrow$)}} \\
		\midrule
        \multirow{3}{*}{sofa} 
        & No Attack &  100.00 & 57.28 & 0.36   \\
		& Camouflage &  100.00\textcolor{red}{\scriptsize{(0.00$\downarrow$)}} & 50.45\textcolor{red}{\scriptsize{(6.83$\downarrow$)}}  & 0.36\textcolor{red}{\scriptsize{(0.00$\uparrow$)}} \\
		& \textit{Ours} Patch &  \textbf{78.57} \textcolor{red}{\scriptsize{(21.43$\downarrow$)}} & \textbf{42.19} \textcolor{red}{\scriptsize{(15.09$\downarrow$)}}  & \textbf{0.52}\textcolor{red}{\scriptsize{(0.16$\uparrow$)}} \\
		\midrule
        \multirow{3}{*}{toilet} 
        & No Attack &  81.25 & 43.32 & 1.32   \\
		& Camouflage &  75.00\textcolor{red}{\scriptsize{(6.25$\downarrow$)}} & 36.85\textcolor{red}{\scriptsize{(6.47$\downarrow$)}}  & 1.49\textcolor{red}{\scriptsize{(0.17$\uparrow$)}} \\
		& \textit{Ours} Patch &  \textbf{56.25} \textcolor{red}{\scriptsize{(25.00$\downarrow$)}} & \textbf{27.42} \textcolor{red}{\scriptsize{(15.90$\downarrow$)}}  & \textbf{1.93}\textcolor{red}{\scriptsize{(0.61$\uparrow$)}} \\
		\midrule
        \multirow{3}{*}{Average} 
        & No Attack &  94.85 & 51.87 & 0.39   \\
		& Camouflage &  84.44\textcolor{red}{\scriptsize{(10.41$\downarrow$)}} & 45.08\textcolor{red}{\scriptsize{(6.79$\downarrow$)}}  & 0.56\textcolor{red}{\scriptsize{(0.17$\uparrow$)}} \\
		& \textit{Ours} Patch &  \textbf{72.46} \textcolor{red}{\scriptsize{(\textbf{22.39}$\downarrow$)}} & \textbf{30.99} \textcolor{red}{\scriptsize{(20.88$\downarrow$)}}  & \textbf{0.89}\textcolor{red}{\scriptsize{(0.50$\uparrow$)}} \\
		\bottomrule
	\end{tabular}
\end{table}

\subsection{Experimental Setup}\label{sec:experimental_setup}

\textbf{Settings.} We use the modular-based navigation agent from~\cite{zhai2023peanut} for ObjectNav, which incorporates a Mask R-CNN~\cite{he2017mask} for object instance segmentation. Our navigation experiment setup follows the 2022 Habitat ObjectNav Challenge~\cite{habitatchallenge2023}. For the adversarial attack, we conduct experiments in ObjectNav using the HM3D dataset~\cite{ramakrishnan2021hm3d} within the Habitat simulator~\cite{habitat19iccv,szot2021habitat}. We perform white-box optimization using scenes from the validation dataset, with attack targets being several categories listed in Table~\ref{tab:main1}.

A $512 \times 512$ adversarial patch with four RGBA channels is used, initialized with Gaussian noise and manually positioned on the target object. Mitsuba 3~\cite{Jakob2020DrJit} serves as the renderer, using planar area lights or \enquote{constant} scene lighting with an intensity parameter of 40., and a camera resolution of $512 \times 512$. During texture and opacity optimization, the detection confidence threshold is set to 0.5. We apply PGD attacks over 100 iterations with a step size of 1.

For each target object category, we randomly select one scene from the HM3D validation set that contains the object and run 100 test episodes with random initial positions. The high success rates in the no-attack condition reflect the strong performance of the modular navigation pipeline under these controlled evaluation settings.

\textbf{Evaluations.} We evaluate attack performance using three metrics from \cite{zhai2023peanut}: Success Rate (SR), Success weighted by Path Length (SPL), and Distance to Goal (DTS). SR measures the agent's ability to successfully locate the target object, while SPL considers both the success rate and the efficiency of the path taken. DTS quantifies the remaining distance between the agent and the goal at the end of each episode. Additionally, we measure the Attack Success Rate (ASR), which is defined as the proportion of viewpoints that were successfully attacked (i.e., viewpoints where detection fails) among all sampled viewpoints.

\subsection{Attack Performance}\label{attack_performance}

\begin{table}[t!]
	\centering
	\scriptsize
	\caption{Attack performance on multi-view object detection.}
	\label{tab:main2}
	\begin{tabular}{c|l}
		\toprule
		Attack Method & ASR$\uparrow$(\%)\\
		\midrule
		\midrule
		No Attack &  19.01   \\
		Camouflage \cite{liu2020spatiotemporal} &  85.12 \textcolor{red}{\scriptsize{(66.11$\uparrow$)}}\\
            Adversarial Patch \scriptsize{(Random Texture)} &  29.75 \textcolor{red}{\scriptsize{(10.74$\uparrow$)}} \\
            Adversarial Patch \scriptsize{(2D Optimized Texture)} &  74.38 \textcolor{red}{\scriptsize{(55.37$\uparrow$)}} \\
		\midrule
		\midrule
		\textit{Ours} Patch \scriptsize{(Multi-view Optimized Texture)}  &  89.26 \textcolor{red}{\scriptsize{(70.25$\uparrow$)}}\\
            \textit{Ours} Patch \scriptsize{(Multi-view Optimized Texture \& Opacity)}& \textbf{98.35}  \textcolor{red}{\scriptsize{(79.34$\uparrow$)}}\\
		\bottomrule
	\end{tabular}
        \vspace{-0.45cm}
\end{table}

\begin{figure}[b!]
        \vspace{-0.45cm}
	\centering
	\includegraphics[width=0.95\linewidth]{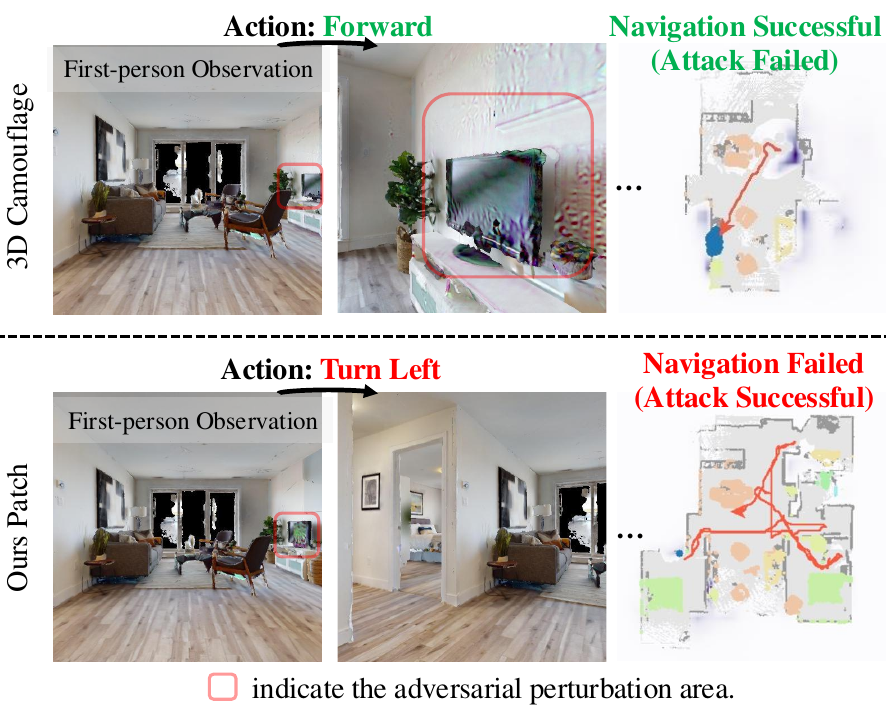}
	\caption{\textbf{Visualization of the attack performance of our adversarial patch compared to 3D camouflage in the navigation task.} The first row shows the adversarial scenario with 3D camouflage, where the agent was still able to successfully navigate and locate the target. In contrast, in the scenario with our adversarial patch, the agent failed to detect the target even after 499 steps, resulting in navigation failure.}
	\label{fig:vis}
\end{figure}

\begin{figure}[h!]
	\centering
	\includegraphics[width=0.95\linewidth]{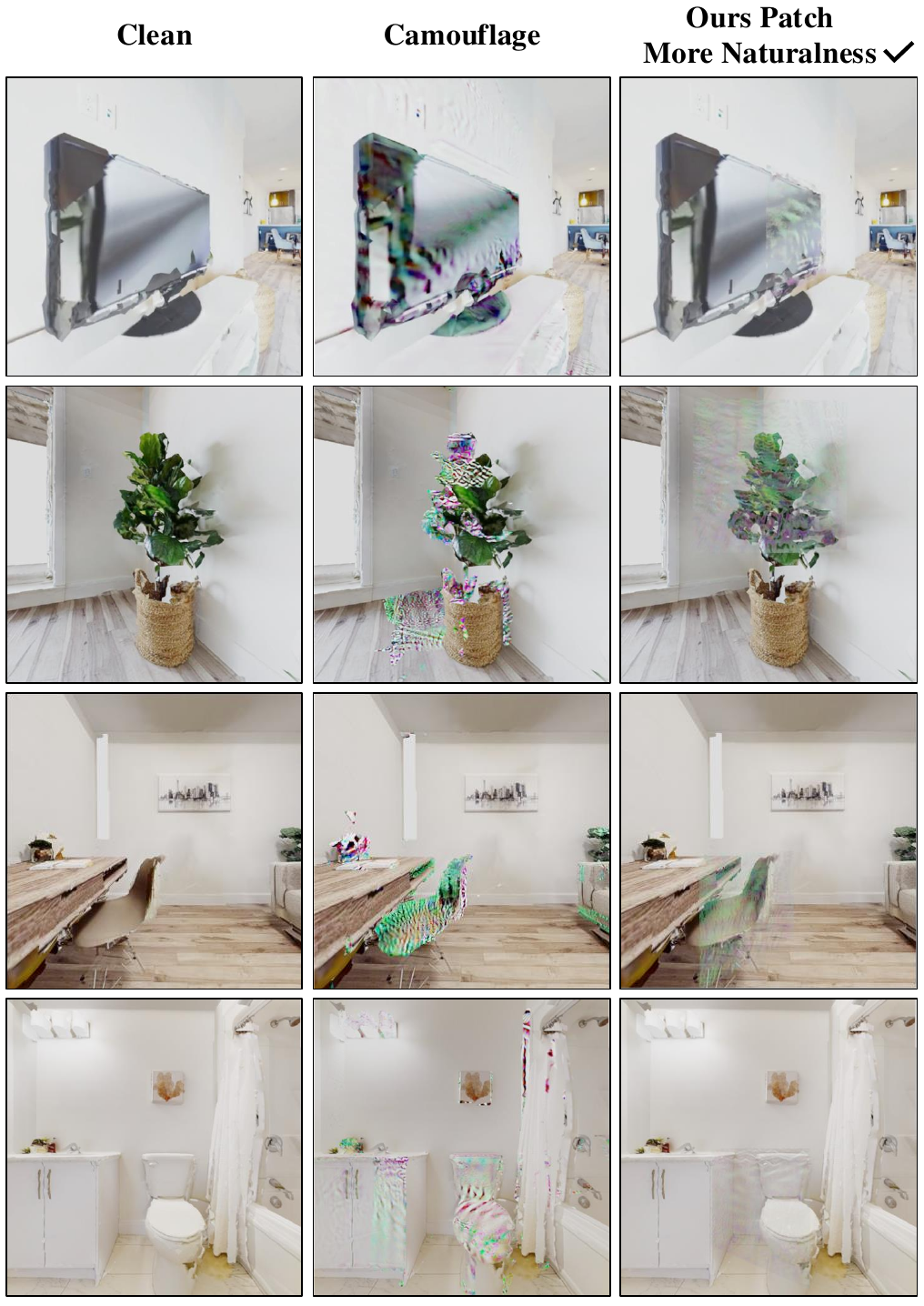}
	\caption{\textbf{Comparison of naturalness.} From left to right, the images show the previous camouflage method, our adversarial patch, and the clean scene. Our adversarial patch exhibits greater visual naturalness.}
	\label{fig:naturalness}
	\vspace{-0.4cm}
\end{figure}

Table~\ref{tab:main1} presents the attack performance of our adversarial patch with multi-view and opacity optimization. As camouflage~\cite{liu2020spatiotemporal} is the only physically realizable attack tailored for embodied navigation, we use it as our baseline. Our method achieves an average SR reduction of 22.39\% across various targets, along with notable decreases in SPL and increases in DTS, indicating less efficient and more difficult navigation. Figure~\ref{fig:vis} visualizes the stronger disruption effect of our approach, while Figure~\ref{fig:naturalness} highlights its superior visual naturalness.

Table~\ref{tab:main2} reports the Attack Success Rate (ASR) for multi-view target detection. We applied our object-aware sampling strategy around a television monitor, generating 120 viewpoints—20 for optimization and 100 unseen for testing. ASR is defined as the proportion of failed detections among these 100 views (confidence threshold 0.5). We compared our method with prior physically realizable attacks, including camouflage~\cite{liu2020spatiotemporal}, 2D-optimized patches without differentiable rendering~\cite{brown2017adversarial, eykholt2018robust, yang2020patchattack, liu2018dpatch, yang2022controllable, kang2024diffender}, and randomly initialized textures. Our method achieved 89.26\% ASR, which further increased to 98.35\% with opacity optimization, clearly outperforming all baselines.

\subsection{Ablation Study}

In this section, we conduct ablation experiments to investigate the impact of two components: the multi-view optimization strategy based on object-aware sampling and the opacity optimization strategy. The objective is to provide insights into the key factors driving performance improvements.

\begin{table}[t!]
	\centering
	\footnotesize
	\caption{Ablation Study on Attack Components.}
	\label{tab:ablation1}
	\setlength{\tabcolsep}{2.9mm}{
		\begin{tabular}{c|c|c|l}
			\toprule
			{\makecell{Texture \\ Optimization}} &  {\makecell{Opacity \\ Optimization}} & Viewpoints & ASR$\uparrow$(\%) \\
			\midrule
			\midrule 
                \xmark & \xmark & - & 19.01\\ 
			\cmark & \xmark & 1 & 74.38 \textcolor{red}{\scriptsize{(55.37$\uparrow$)}} \\
                \cmark & \xmark & 20 & 89.26 \textcolor{red}{\scriptsize{(70.25$\uparrow$)}} \\
			\cmark & \cmark & 20 & \textbf{98.35}  \textcolor{red}{\scriptsize{(79.34$\uparrow$)}} \\
			\bottomrule
	\end{tabular}}
        \vspace{-0.45cm}
\end{table}

\textbf{Effect of multi-view optimization based on object-aware sampling.} The experimental results are shown in Table~\ref{tab:ablation1}. As observed, when neither multi-view texture optimization nor opacity optimization is applied, the ASR is the lowest at 19.01\%. When only single-view texture optimization is applied, there is an improvement, and multi-view texture optimization leads to even higher performance. When both texture and opacity are optimized, the ASR reaches its highest value of 98.35\%, validating the effectiveness of our multi-view optimization strategy.

\begin{figure}[bh!]
	\centering
	\includegraphics[width=0.95\linewidth]{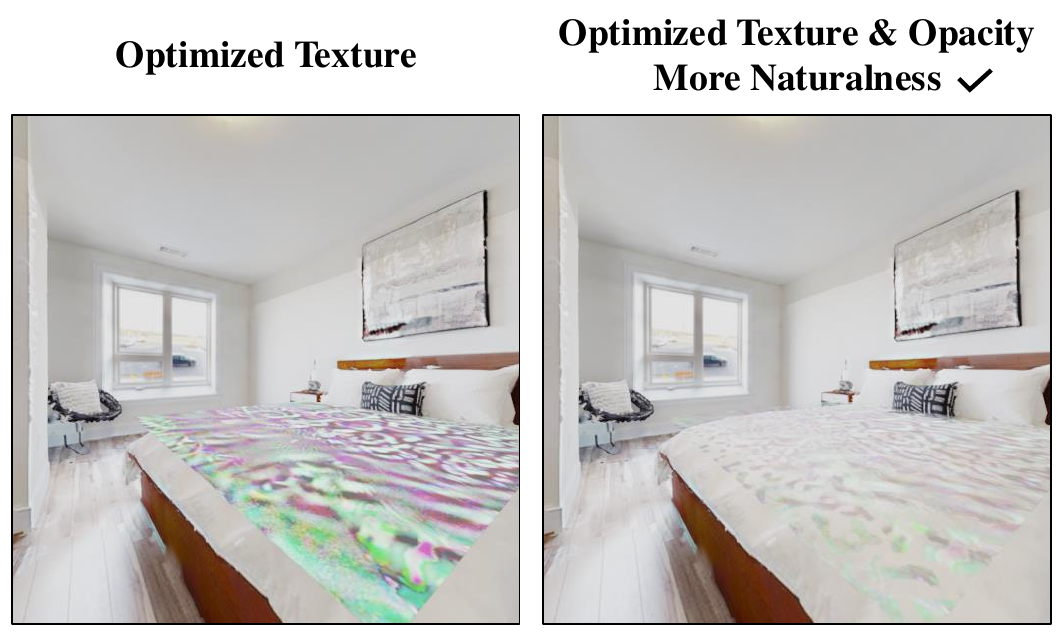}
	\caption{\textbf{Visualization of the effect of opacity optimization.} It is evident that after applying opacity optimization, the naturalness of the patch improves significantly.}
	\label{fig:ablation}
\end{figure}

\begin{figure}[t!]
	\centering
	\includegraphics[width=0.95\linewidth]{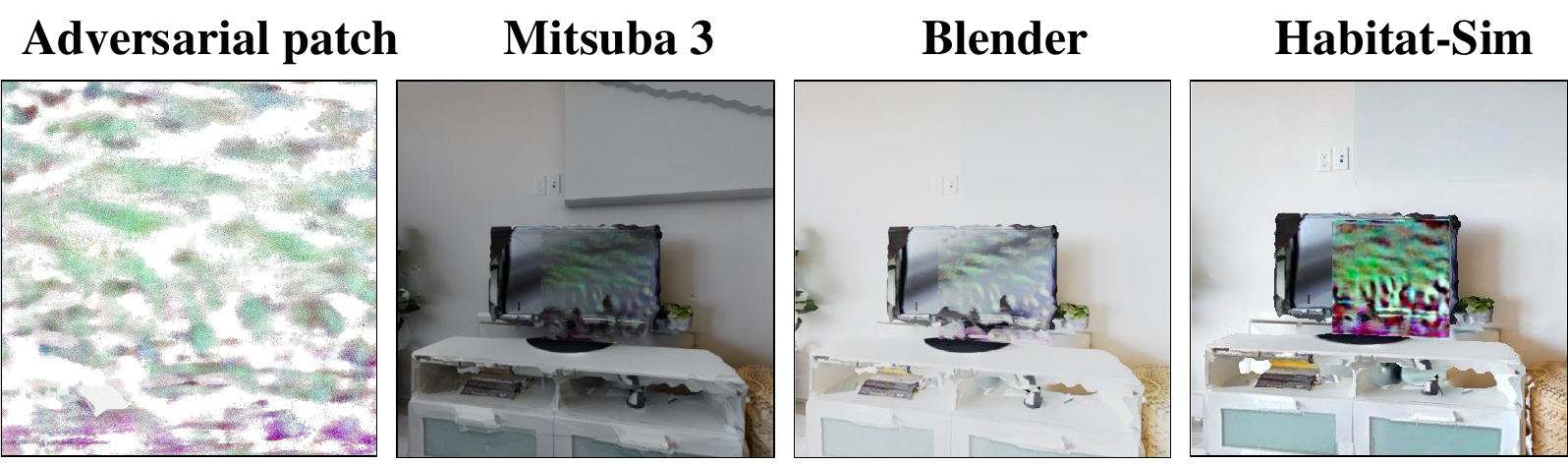}
	\caption{\textbf{Visualization of adversarial patches with opacity under different renderers.} Due to differences in physical rendering techniques and lighting setups, this highlights the challenge of ensuring cross-renderer transferability and maintaining attack effectiveness. Notably, the habitat-sim renderer lacks support for transparent materials, making the texture appear more visible in its renderings.}
	\label{fig:ablation3}
\end{figure}

\textbf{Effect of opacity optimization.} As shown in Figure~\ref{fig:ablation}, opacity optimization significantly improves the patch’s visual naturalness. We further examined different initial opacity levels (Table~\ref{tab:ablation2}) on both random and optimized textures. Lower opacity (e.g., 0.2) renders the patch nearly invisible, while higher values (e.g., 0.8) improve ASR but reduce stealth. This positive correlation mainly holds for optimized textures; random textures lack alignment with model vulnerabilities, making opacity less effective. We selected an initial opacity of 0.6 to balance attack strength and visibility, and then refined the mask accordingly. Figure~\ref{fig:ablation3} shows the patch's appearance under different renderers.

Physically, the patch can be printed on transparent paper with a laser printer and directly affixed to target objects, avoiding any modification to the scene geometry. Its adjustable transparency further enhances stealth in real-world navigation scenarios.

\begin{table}[t!]
	\centering
	\footnotesize
	\caption{Ablation Study on Opacity Value.}
	\label{tab:ablation2}
	\setlength{\tabcolsep}{2.9mm}{
		\begin{tabular}{c|c|l}
			\toprule
			Method & Opacity Value(\%) & ASR$\uparrow$(\%) \\
			\midrule
			\midrule 
                \multirow{5}{*}{Texture Random} 
                & 0.20 & 24.79 \\
			& 0.40 & 21.49 \\
                & 0.60 & 15.70 \\
                & 0.80 & 29.75 \\
                \midrule
			\midrule
                \multirow{5}{*}{Texture Optimized} 
                & 0.20 & 22.31 \\
			& 0.40 & 71.90 \\
                & \textbf{0.60} & \textbf{90.91} \\
                & 0.80 & 87.60 \\
			\bottomrule
	\end{tabular}}
	\vspace{0.45cm}
\end{table}

\section{Conclusions}
In this paper, we propose a practical attack on embodied navigation by applying adversarial patches with learnable textures and opacity to objects, reducing success rates by 22.39\% on average. While the method shows strong performance in simulation, future work is needed to improve patch printing techniques and evaluate human perceptibility in real-world settings. Despite the lack of physical experiments, our approach outperforms prior methods—including first-person perturbations, full 3D texture attacks, and less effective multi-view patches—in physical feasibility, multi-view robustness, and visual naturalness. Moreover, Virtual 3D environments continue to play a key role in advancing embodied navigation research.

\section*{Acknowledgment}
This work was supported by the Beijing Natural Science Foundation (Grant No. F2024203115), the National Natural Science Foundation of China (Grant No. 62173045), the National Natural Science Foundation of China (Grant No. 62403491), the China Postdoctoral Science Foundation (Grant No. 2024M750255), the \enquote{Double First-Class} Interdisciplinary Team Project of Beijing University of Posts and Telecommunications (Grant No. 2023SYLTD02), and the National Key R\&D Program of China (Grant No. 2024YFC3015604).

\bibliographystyle{IEEEtran}
\bibliography{root}

\end{document}